\documentclass[conference]{IEEEtran}
\IEEEoverridecommandlockouts
\usepackage{cite}
\usepackage{amsmath,amssymb,amsfonts}
\usepackage{algorithmic}
\usepackage{graphicx}
\usepackage{textcomp}
\usepackage{xcolor}
\usepackage{booktabs}
\usepackage{adjustbox}
\usepackage{array}
\usepackage{multicol}
\usepackage{multirow}
\usepackage{makecell}
\def\BibTeX{{\rm B\kern-.05em{\sc i\kern-.025em b}\kern-.08em
    T\kern-.1667em\lower.7ex\hbox{E}\kern-.125emX}}
\begin{document}

\title{TongueSAM: An Universal Tongue Segmentation Model Based on SAM with Zero-Shot\\
}
\author{\IEEEauthorblockN{Shan Cao}
\IEEEauthorblockA{\textit{Xiamen University} \\
Xiamen, China \\
30920201153954@stu.xmu.edu.cn}

\and
\IEEEauthorblockN{Qingfeng Wu*}
\IEEEauthorblockA{\textit{Xiamen University}\\ 
Xiamen, China \\
*Correspondence Author: qfwu@xmu.edu.cn}
\and
\IEEEauthorblockN{Linjian Ma}
\IEEEauthorblockA{\textit{Xiamen University}\\ 
Xiamen, China \\
malinjian@xmu.edu.cn}
}

\maketitle

\begin{abstract}
Tongue segmentation serves as the primary step in automated TCM tongue diagnosis, which plays a significant role in the diagnostic results. Currently, numerous deep learning based methods have achieved promising results. However, when confronted with tongue images that differ from the training set or possess challenging backgrounds, these methods demonstrate limited performance. To address this issue, this paper proposes a universal tongue segmentation model named TongueSAM based on SAM (Segment Anything Model). SAM is a large-scale pretrained interactive segmentation model known for its powerful zero-shot generalization capability. Applying SAM to tongue segmentation leverages its learned prior knowledge from natural images, enabling the achievement of zero-shot segmentation for various types of tongue images. In this study, a Prompt Generator based on object detection is integrated into SAM to enable an end-to-end automated tongue segmentation method. Experiments demonstrate that TongueSAM achieves exceptional performance across various of tongue segmentation datasets, particularly under zero-shot. Even when dealing with challenging background tongue images, TongueSAM achieves a mIoU of 95.23\% under zero-shot conditions, surpassing other segmentation methods. As far as we know, this is the first application of large-scale pretrained model for tongue segmentation. The project mentioned in this paper is currently publicly available\footnote{https://github.com/cshan-github/TongueSAM}.

\end{abstract}


\section{Introduction}
Tongue segmentation refers to extract the tongue region from facial or oral images of patients and is a critical step in automated Traditional Chinese Medicine (TCM) tongue diagnosis\cite{1}. The task of tongue segmentation presents certain difficulties due to the rich morphological and color features of the tongues, as well as the similarity in color between organs(lips,chin) and tongues. Several researches have employed deep neural network(DNN) based methods to address tongue segmentation problems and achieved remarkable experimental results \cite{2,3,4,5,6}. However, these methods exhibit limited performance on datasets with distributions different from the training set\cite{7}. This is largely due to the lack of large-scale tongue segmentation datasets, which results in pretrained models lacking applicability in changing scenarios. Meanwhile, it is hard to directly transfer models pretrained on other domain datasets to the task of tongue segmentation due to its unique nature. Consequently, many automated TCM tongue diagnosis methods still rely on color thresholding and contour-based segmentation methods for tongue segmentation\cite{8,9,10}, leaving room for improvement in segmentation accuracy. In a recent study by \cite{7}, a training framework called "Iterative cross-domain tongue segmentation" was proposed. This framework aims to improve the performance of the model in the target domain by fine-tuning the model using samples from the target domain. However, this approach requires obtaining data from the target domain in advance, which can impose limitations in real-time. Therefore, this paper aim to develop a universal segmentation model that without additional training. 

Recently, the "Segmentation Anything Model" (SAM) has gained widespread attention\cite{11} due to its versatile segmentation capabilities. Applying SAM to tongue segmentation can effectively leverage its strong zero-shot transfer capability. However, as an interactive segmentation model, the quality of segmentation results have related to manual prompts. Although SAM supports segmentation without prompts, prompt-free SAM performs lacks the advantages of segmentation accuracy and robustness. We discovered that by utilizing object detection methods, it is possible to generate box prompts that closely resemble the ones provided manually. This allows the SAM to achieve end-to-end high-quality segmentation without manual intervention. Based on the research background mentioned above, this paper proposes a tongue segmentation method called TongueSAM based on SAM, with the addition of an object detection based Prompt Generator. In the zero-shot scenario, TongueSAM demonstrates significantly performance compared to other tongue segmentation methods.

\section{Related Work}

Tongue segmentation, as a medical image segmentation task, has received considerable attention. Over the past few years, numerous segmentation methods have utilized low-level features such as color and texture in tongues. In \cite{12}, a method combining dual elliptical deformable templates and active contour models was proposed, demonstrating improved stability and accuracy in tongue segmentation. Another approach proposed in \cite{13} employed a combination of polar coordinate edge detectors and active contour models to achieve accurate segmentation of tongue contours. Additionally,\cite{14} introduced a novel color tongue image segmentation algorithm based on the HSI model, which exhibited faster speed and superior performance.

With the rapid development of deep neural networks, increasing researches focused on leveraging deep semantic segmentation methods for tongue segmentation. \cite{2} presented an end-to-end model called TongueNet, which employed multi-task learning for tongue segmentation. This model effectively leverages pixel-level prior information to achieve superior performance in tongue segmentation. Furthermore, \cite{3} proposed a Dilated Encoder Network (DE-Net) that captured more advanced features and attained high-resolution outputs, outperforming contemporary methods in tongue segmentation. \cite{4} introduced DeepTongue, significantly improved segmentation precision and efficiency for tongue segmentation without the need for preprocessing. In \cite{5}, an enhanced tongue segmentation method based on Deeplabv3+ and a loss function based on edge information was proposed, which improved the network's ability to extract multi-scale and low-level information and encouraged the model to focus on separating tongue boundaries during training. Additionally, \cite{6} presented a deep semantic enhancement network named DSE-Net, composed of lightweight feature extraction modules, efficient deep semantic enhancement modules, and decoders. Experimental results demonstrated that DSE-Net outperformed other mobile tongue segmentation methods on two different datasets.

The Segment Anything Model (SAM) has been proposed as a benchmark model for natural image segmentation \cite{11}, thanks to being trained on over 1.1 million natural images with over 1 billion ground truth segmentation masks. With the advent of SAM, researchers have explored its performance in various domains, with medical image segmentation garnering significant attention. In \cite{15}, SAM was successfully extended to medical image segmentation tasks, including adaptation of empirical benchmarks and methodologies, while discussing potential future directions of SAM in medical image segmentation. \cite{16} explored SAM's zero-shot performance in medical image segmentation by implementing eight different cueing strategies on six datasets from four imaging modalities. The study demonstrated that SAM's zero-shot performance was comparable to, and in some cases even surpassed, current state-of-the-art methods. \cite{17} designed SkinSAM, a model refined through training based on SAM, which exhibited outstanding segmentation performance in skin cancer segmentation. \cite{18} reported quantitative and qualitative zero-shot segmentation results on nine medical image segmentation benchmarks. These benchmarks encompassed various imaging modalities and different application domains. Experimental results revealed that SAM's zero-shot segmentation performance in medical images remained limited. \cite{19} extensively evaluated SAM on 19 medical image datasets from multiple modalities and anatomical structures, drawing the conclusion that SAM exhibited impressive zero-shot segmentation performance on certain medical image datasets, but moderate to poor performance on others. Some recent studies suggest that although SAM has achieved remarkable results in natural image segmentation, there is still room for improvement in medical image segmentation\cite{18,19}.

\begin{figure*}[!ht]
\centering
\includegraphics[width=110mm]{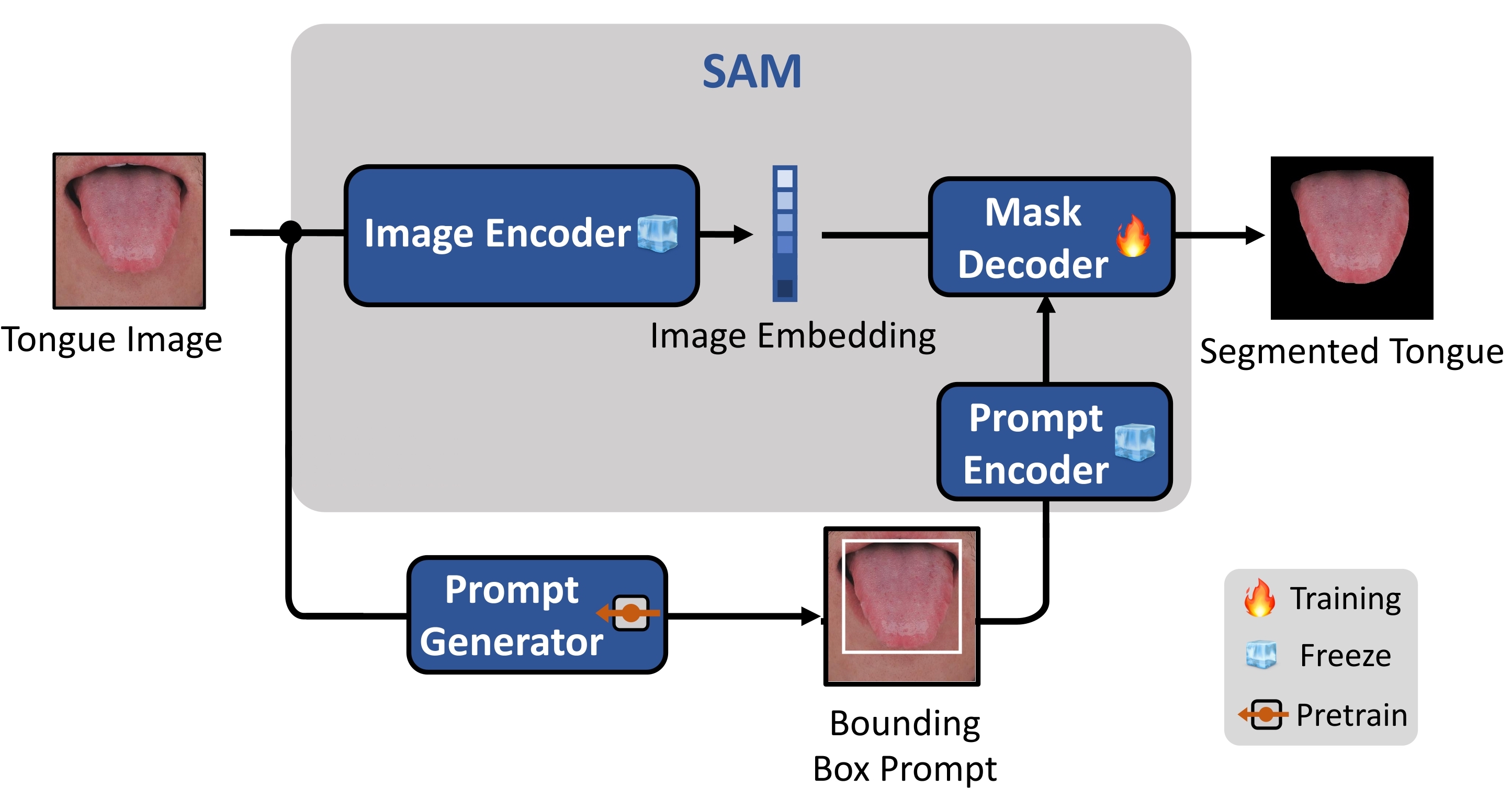}
\caption{The overall framework of TongueSAM.}
\end{figure*}

\section{Method}
The proposed TongueSAM architecture as shown in the Fig. 1. TongueSAM consists primarily of two components: SAM and the Prompt Generator. For a given tongue image, TongueSAM first utilizes the pretrained Image Encoder in SAM for encoding. Meanwhile, the Prompt Generator generates bounding box prompt based on the tongue image. Finally, the image embedding and prompts are jointly fed into the Mask Decoder to generate the segmentation result. The entire segmentation process is end-to-end and does not require any additional manual prompts. The following sections will introduce different components of TongueSAM.

\subsection{SAM}
In order to leverage the distinctive features of the pretrained SAM model, this paper strives to preserve the original model structure of SAM. The overall SAM model is built upon the Transformer architecture and consists of three components: Image Encoder, Prompt Encoder, and Mask Decoder. The structure of each component will be descripted as follows.

{
\setlength{\parindent}{0cm}
\textbf{1.Image Encoder}
}

SAM utilizes a Vision Transformer (ViT) pretrained with MAE \cite{20} as Image Encoder. Specifically, SAM employs ViT-H/16, which consists of 14*14 window attentions and four equidistant global attention blocks. The output of the image encoder is a 16* downsampled embedding of the input image. Regarding input dimensions, SAM operates with an input resolution of 1024*1024, resulting in a final image embedding size of 64*64. Additionally, to reduce channel dimensions, a 1*1 convolutional layer is employed to decrease the number of channels to 256. Subsequently, a 3*3 convolutional layer with 256 channels is used, followed by layer normalization after each convolutional layer. 

{
\setlength{\parindent}{0cm}
\textbf{2.Prompt Encoder}
}

The Prompt Encoder of SAM accepts two types of prompts: sparse prompts (points, boxes, text) and dense prompts (masks). Sparse prompts, such as points and boxes, are represented in the form of positional encodings and added to the learned embeddings of each prompt type. Free-form text prompts are encoded using a readily available text encoder from CLIP\cite{21}. Dense prompts (masks) are embedded using convolutions and element-wise summed with the image embedding. The Prompt Encoder of SAM is capable of mapping user-provided sparse prompts to 256-dimensional vectorized embeddings. A point is represented by the positional encoding of its location combined with two learned embeddings, indicating whether the point belongs to the foreground or background. A box is represented by a pair of embeddings representing the positional encodings of the top-left and bottom-right corners. Dense prompts (masks) have a spatial correspondence with the image. SAM takes in masks at a resolution 4 times lower than the input image, and then applies two 2*2 convolutions with a stride of 2, yielding output channels of 4 and 16, respectively, resulting in a further downsampling by 4. Finally, a 1*1 convolutional layer maps the channel dimension to 256. Each layer is followed by GELU activation and layer normalization. The mask embedding and the image embedding are then element-wise added. 

{
\setlength{\parindent}{0cm}
\textbf{3.Mask Decoder}
}

The Mask Decoder efficiently maps the image embedding and a set of prompt embeddings to output masks. To combine these inputs, SAM draws inspiration from Transformer segmentation models\cite{22,23} and modifies the standard Transformer decoder. Before applying the decoder, a learned output token embedding is inserted into the set of prompt embeddings, which will be used for decoding.

The decoder in SAM is designed as shown in Fig. 2. Each decoder layer performs four steps: (1) self-attention over tokens, (2) cross-attention from tokens (as queries) to image embedding, (3) point-wise MLP updates to each token, and (4) cross-attention from image embedding (as queries) to tokens. The last step uses prompt information to update the image embedding. During cross-attention, the image embedding is treated as a set of 64*64 256-dimensional vectors. Each self/cross-attention and MLP has residual connections, layer normalization, and a dropout rate of 0.1 during training. The next decoder layer utilizes the updated tokens and image embedding from the previous layer. SAM employs two decoder layers in total.

\begin{figure}[!ht]
\centering
\includegraphics[width=80mm]{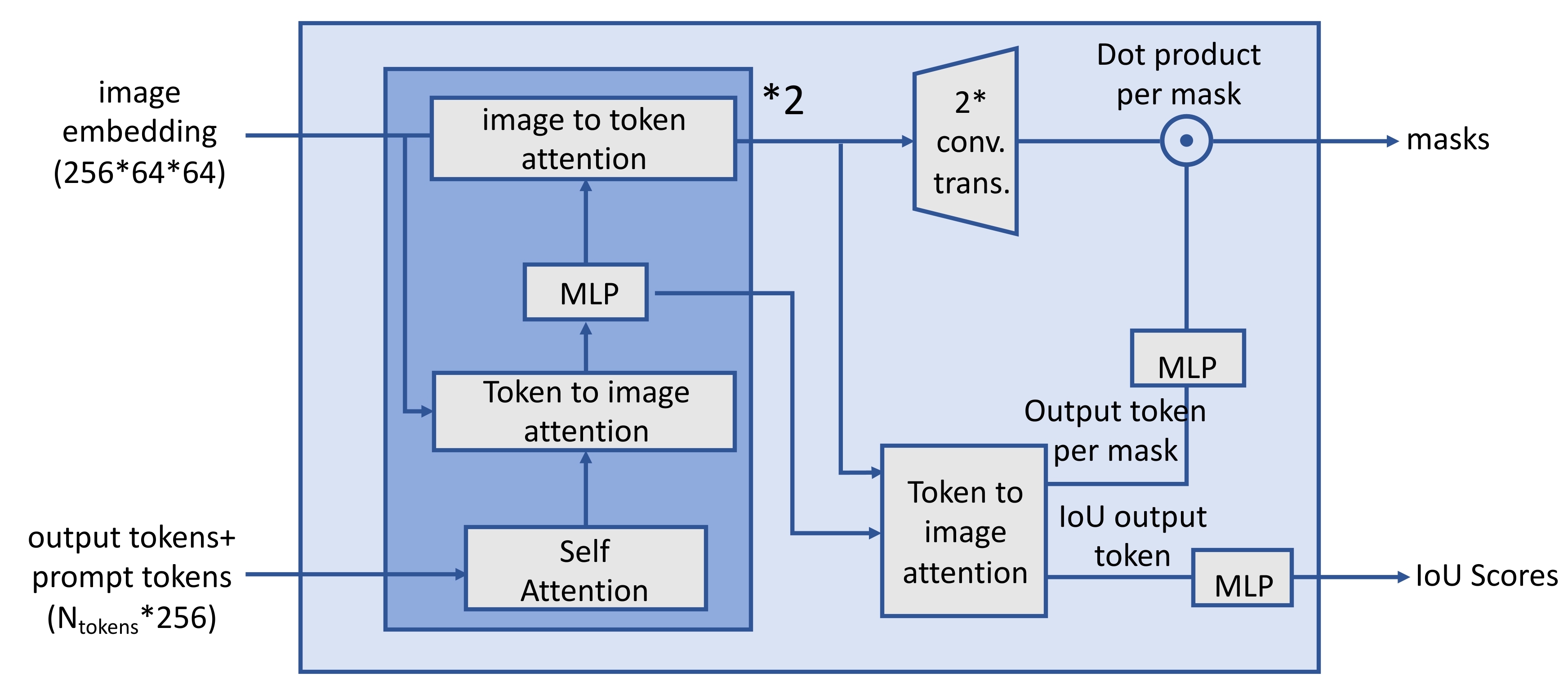}
\caption{Details of the Mask Decoder\cite{11}. The dual-layer decoder updates the image embeddings and prompt tokens through cross-attention. Subsequently, the image embeddings are amplified, and the updated outputs are used to dynamically predict masks. In each attention layer, positional encoding is added to the image embeddings, and the entire original prompt token is readded to the token queries and keys.}
\end{figure}

\subsection{Prompt Generator}

Due to SAM being an interactive segmentation model, the manual prompts has a significant impact on the segmentation results. In this study, different types of prompts were compared in section 4.1. The results indicate that, bounding box prompts yield the best segmentation performance in tongue segmentation, which aligns with conclusions from other studies \cite{15,16,17,18,19}.

To automate the entire segmentation process, this study proposes a Prompt Generator based on object detection. Object detection can identify specific objects in an image, which precisely with the bounding box prompt supported by SAM. This approach effectively improves the performance of SAM in the absence of prompts and achieves results close to those provided by ground truth with bounding boxes. 

\subsection{TongueSAM}

To preserve the pretrained weights in SAM, this study employs different training strategy for the components in TongueSAM. The Prompt Generator is trained  as object detection methods in advance and is not involved in the overall model's backpropagation. The Image Encoder and Mask Encoder contain a significant amount of pretrained weights from natural images. Training these components would require a substantial amount of training data and computational resources. Therefore, we keep them frozen during the training process. The Mask Decoder plays a crucial role in determining the final segmentation results and is a lightweight transformer decoder. It needs to be fine-tuned specifically for the task of tongue segmentation.

\begin{figure*}[!ht]
\centering
\includegraphics[width=150mm]{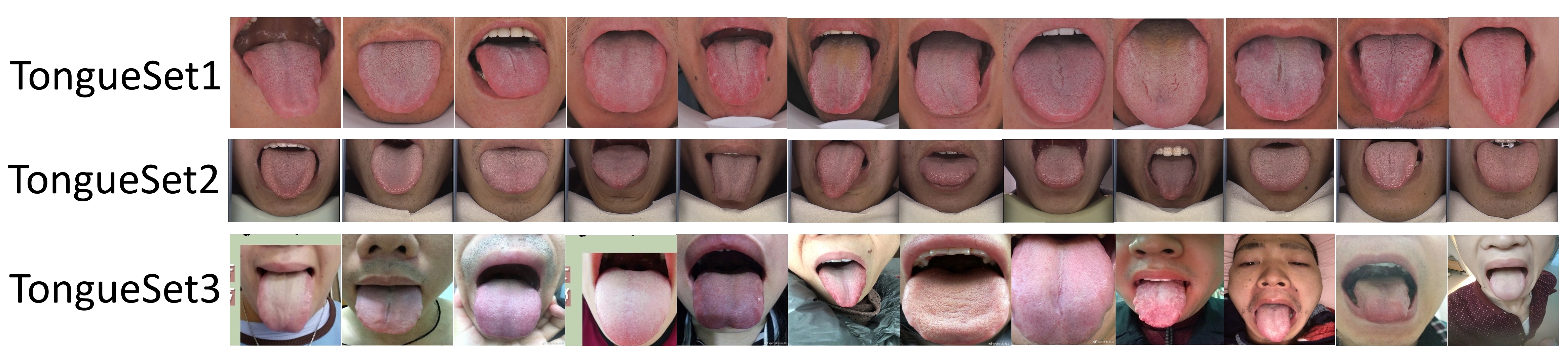}
\caption{Samples of TongueSet1, TongueSet2 and TongueSet3.}
\end{figure*}

\section{Experiment}

To thoroughly validate the effectiveness of TongueSAM, experiments were conducted on three datasets, the samples of these datasets are shown as Fig. 3. The first dataset, called TongueSet1, comprised 3192 tongue images gathered using TFDA-1 tongue diagnostic instruments in collaboration with Shanghai University of Traditional Chinese Medicine. Each patient's tongue image was captured under standardized environmental and manual labeled using the labelme tool. The size of each tongue image were 400*400 pixels. This dataset, characterized by a relatively large amount of data and standardized capture conditions, served as the benchmark pretraining dataset for testing the model's transferability.

The second dataset, called BioHit, was a widely-used public dataset for tongue segmentation\cite{24}. BioHit was introduced by Harbin Institute of Technology and comprised 300 RGB tongue images, along with corresponding manual masks as ground truth. All tongue images were acquired in a semi-enclosed environment under stable lighting conditions, with the image size of 768*576 pixels. It was denoted as TongueSet2 in this study.

The third dataset was obtained from the webset\cite{25}. It encompassed a diverse range of tongue images captured using various sources. A total of 1000 tongue images with varying sizes from this dataset were manual annotated using the labelme tool. This dataset effectively simulated tongue image segmentation tasks in real-world environments, enabling testing of the model's robustness and adaptability.

In this study, we utilized three evaluation metrics to assess the performance of the model: mean Intersection over Union (mIoU), mean Pixel Accuracy (mPA), and Accuracy (Acc). 



The hardware device cpu used in the experiment is Intel(R) Core(TM) i9-7940X, its memory is 16G, the gpu we used is GeForce RTX 2080 Ti. All the experiments were implemented using the PyTorch framework.

\subsection{Comparison of different prompt for SAM tongue segmentation}

This paper focuses on comparing the performance of SAM in tongue segmentation using point prompts, box prompts, and no prompt. The generation of point prompts involves randomly selecting  points from the ground truth, while for box prompts, the left-top and right-top coordinates of the ground truth are used. TongueSet1 is used as the training set, while TongueSet2 and TongueSet3 are used as the test sets. This experimental setup allows for a more refined observation of how prompts affect model transferability. The metric observed is mIoU, and the trends in mPA and Acc changes are similar to mIoU. In terms of training parameters, we choose the optimization of Adam, the learning rate is $1*10^{-4}$, the training epoch is 20, the batch size is 32,and the best on the test set was selected as the final model. The experimental results are presented in Fig. 4.

\begin{table*}      
\caption{Comparison of the different Prompt Generators. The best results are highlighted in bold, while the second-best results are indicated with an underline.}  
\begin{adjustbox}{center}
\renewcommand\arraystretch{1}  
\begin{tabular}{|c|c|c|c|c|c|c|c|c|c|c|}
\hline   &\multirow{2}{*}{}&\multicolumn{3}{c|}{TongueSet1}&\multicolumn{3}{c|}{TongueSet2}&\multicolumn{3}{c|}{TongueSet3} \\   
\cline{3-11}&&mIoU&mPA&Acc&mIoU&mPA&Acc&mIoU&mPA&Acc \\   
\hline   \multirow{4}{*}{Object }&YOLOV3&\underline{94.69\%}&97.16\%&\underline{97.31\%}&93.45\%&\underline{97.41\%}&97.22\%&\underline{92.50\%}&\underline{96.20\%}&\underline{96.20\%}  \\ 
\cline{2-11}  \multirow{4}{*}{Detection}&YOLOV5&93.82\%&96.75\%&96.85\%&\underline{93.64\%}&97.11\%&\underline{97.32\%}&90.95\%&95.46\%&95.37\%  \\
\cline{2-11}  &YOLOX&94.68\%&\underline{97.24\%}&97.29\%&\textbf{94.25\%}&\textbf{97.52\%}&\textbf{97.58\%}&\textbf{92.66\%}&\textbf{96.24\%}&\textbf{96.29\%}  \\ 
\cline{2-11}  &YOLOV7&\textbf{95.11\%}&\textbf{97.39\%}&\textbf{97.53\%}&92.95\%&97.18\%&96.99\%&92.00\%&95.88\%&95.92\%  \\ 
\cline{2-11}  &Faster-RCNN&92.08\%&96.25\%&96.64\%&91.89\%&95.75\%&95.82\%&90.14\%&94.71\%&94.95\%  \\ 
\hline  \multirow{2}{*}{Semantic }&UNet&90.09\%&94.42\%&94.89\%&44.50\%&72.30\%&61.73\%&65.45\%&81.75\%&79.13\%  \\ 
\cline{2-11}  \multirow{2}{*}{Segmentation }&Deeplabv3&93.50\%&96.45\%&96.69\%&70.89\%&87.98\%&84.27\%&75.21\%&87.58\%&85.88\%  \\ 
\cline{2-11}  &PSPNet&93.59\%&96.70\%&96.72\%&75.36\%&89.55\%&87.39\%&81.67\%&90.70\%&90.03\%  \\ 
\hline   
\end{tabular}   
\end{adjustbox}
\end{table*}

\begin{figure}[!ht]
\centering
\includegraphics[width=90mm]{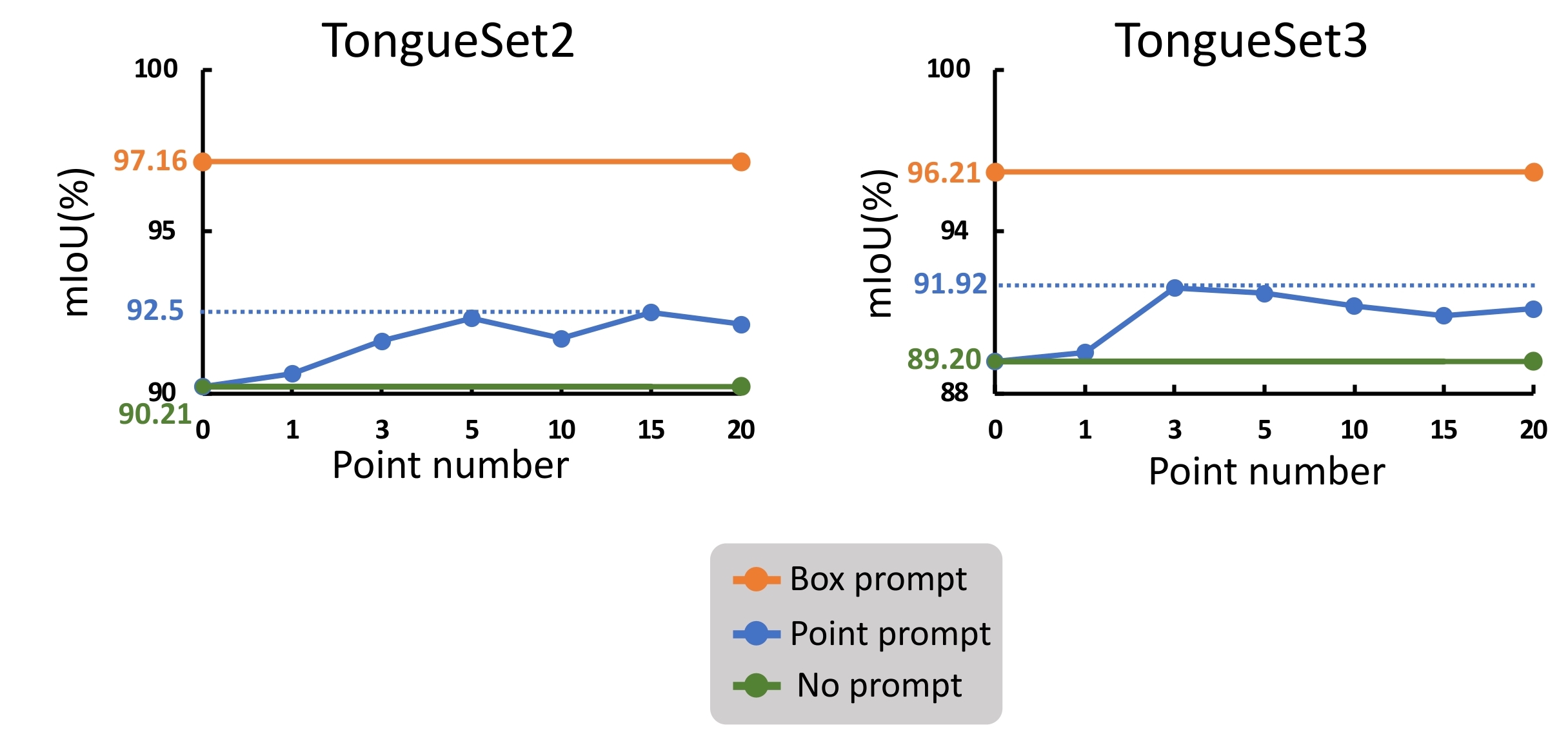}
\caption{Comparison of different prompts for SAM.}
\end{figure}

\begin{table*}[htb]   
\caption{Comparison of the different tongue segmentation methods. The best results are highlighted in bold, while the second-best results are indicated with an underline.}  
\begin{adjustbox}{center}
\renewcommand\arraystretch{1}  
\begin{tabular}{|c|c|c|c|c|c|c|c|c|c|}
\hline   \multirow{2}{*}{}&\multicolumn{3}{c|}{TongueSet1}&\multicolumn{3}{c|}{TongueSet2}&\multicolumn{3}{c|}{TongueSet3} \\   
\cline{2-10}&mIoU&mPA&Acc&mIoU&mPA&Acc&mIoU&mPA&Acc \\   
\hline   Attn.Net&96.41\%&\underline{98.23\%}&98.19\%&89.44\%&93.23\%&95.99\%&94.77\%&97.48\%&97.60\%  \\ 
\hline Deeplabv3&95.83\%&98.02\%&97.90\%&89.41\%&92.99\%&96.01\%&97.53\%&98.79\%&\underline{98.89\%}  \\
\hline FCN&95.50\%&97.72\%&97.73\%&86.82\%&91.46\%&94.93\%&86.07\%&93.10\%&93.22\%  \\ 
\hline PSPNet&95.37\%&97.63\%&97.66\%&89.48\%&93.14\%&96.01\%&95.42\%&97.80\%&97.91\%  \\ 
\hline  U$^2$Net&96.44\%&98.25\%&98.21\%&89.67\%&93.10\%&96.10\%&97.20\%&98.48\%&98.74\%  \\ 
\hline  UNet&95.43\%&97.83\%&97.69\%&88.27\%&92.23\%&95.54\%&96.53\%&98.27\%&98.43\%  \\ 
\hline  SAM without prompt&96.40\%&98.18\%&98.19\%&98.34\%&99.28\%&99.45\%&97.31\%&98.64\%&98.77\%  \\
\hline  SAM with GT box&\textbf{96.98\%}&\textbf{98.46\%}&\textbf{98.49\%}&\textbf{98.62\%}&\textbf{99.41\%}&\textbf{99.55\%}&\textbf{97.85\%}&\textbf{99.00\%}&\textbf{99.02\%}  \\
\hline   
TongueSAM&\underline{96.51\%
}&\underline{98.23\%}&\underline{98.25\%}&\underline{98.41\%}&\underline{99.38\%}&\underline{99.48\%}&\underline{97.55\%}&\underline{98.83\%}&98.88\%  \\
\hline   
\end{tabular}   
\end{adjustbox}
\end{table*}

\begin{table*}[htb]   
\caption{Comparison of the different tongue segmentation methods with zero-shot. The best results are highlighted in bold, while the second-best results are indicated with an underline.}  
\begin{adjustbox}{center}
\renewcommand\arraystretch{1}  
\begin{tabular}{|c|c|c|c|c|c|c|}
\hline   \multirow{2}{*}{}&\multicolumn{3}{c|}{TongueSet2}&\multicolumn{3}{c|}{TongueSet3} \\   
\cline{2-7}&mIoU&mPA&Acc&mIoU&mPA&Acc \\   
\hline   Attn.Net&91.60\%&95.85\%&96.88\%&86.85\%&94.36\%&93.48\%  \\ 
\hline Deeplabv3&90.76\%&96.14\%&96.49\%&92.37\%&96.69\%&96.39\% \\
\hline FCN&92.35\%&95.99\%&97.19\%&89.43\%&95.13\%&94.90\%\\ 
\hline PSPNet&90.76\%&95.21\%&96.56\%&91.56\%&95.65\%&96.03\%\\ 
\hline  U$^2$Net&92.77\%&96.45\%&97.34\%&93.61\%&97.00\%&97.02\%\\
\hline  UNet&85.06\%&93.78\%&93.98\%&87.41\%&93.93\%&93.86\%\\
\hline  UNet++&84.84\%&94.51\%&93.77\%&82.62\%&92.29\%&91.08\%\\ 
\hline  SAM without prompt&90.21\%&95.99\%&96.27\%&89.20\%&94.58\%&94.83\%\\
\hline  SAM with GT box&\textbf{97.16\%}&\underline{98.16\%}&\textbf{99.00\%}&\textbf{96.77\%}&\textbf{98.24\%}&\textbf{98.53\%}\\
\hline  TongueSAM&\underline{96.24\%}&\textbf{98.39\%}&\underline{98.74\%}&\underline{95.23\%}&\underline{97.68\%}&\underline{97.87\%}\\
\hline
\end{tabular}   
\end{adjustbox}
\end{table*}

\begin{figure*}[!ht]
\centering
\includegraphics[width=180mm]{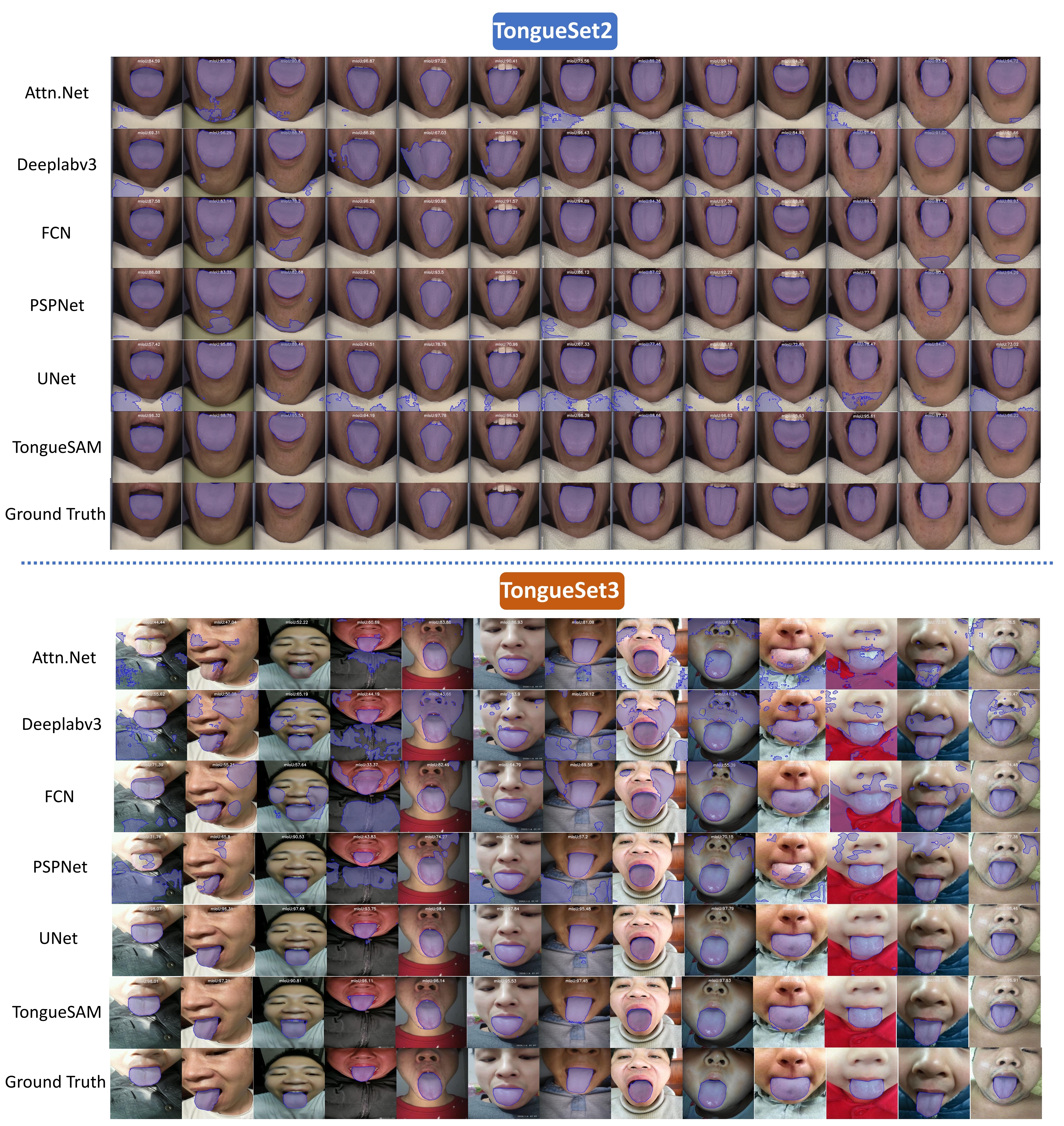}
\caption{Samples of segmentation using different methods with zero-shot. The blue range represents the segment area of the model.}
\end{figure*}

The experimental results indicate that using box prompts in SAM yields favorable performance on both TongueSet2 and TongueSet3. On the other hand, when employing point prompts in SAM, increasing the number of points beyond 5 does not significantly improve the performance. Inspired by these results, this study utilizes object detection methods to automatically generate box prompts. To identify the most suitable object detection method, several commonly used object detection and semantic segmentation methods(The left-top and right-top coordinates of the masks are used as the box prompts) are compared in terms of the quality of generated box prompts. In this study, 80\% of TongueSet1 is selected as the training set, while the remaining 20\% of TongueSet1, TongueSet2, and TongueSet3 are used for testing. The hyperparameters of the object detection methods are optimization of Adam, the learning rate is $1*10^{-4}$, the training epoch is 500, the batch size is 16. The hyperparameters of the semantic segmentation methods are optimization of Adam, the learning rate is $1*10^{-4}$, the training epoch is 100, the batch size is 8. The result is shown in Table 1.

The experimental results indicate that the object detection methods far surpasses semantic segmentation methods, largely due to their distinct training tasks. Furthermore, object detection methods demonstrate excellent performance across various datasets with zero-shot. YOLOX is selected as the Prompt Generator for TongueSAM in this study due to its notable detection capability, which has shown to be relatively superior across diverse datasets.

\subsection{Comparison with other tongue segmentation methods.}

To validate the effectiveness of TongueSAM, this study compares TongueSAM with several commonly used segmentation methods on three datasets separately. For each dataset, 80\% of the data is used as the training set, and the remaining 20\% is used as the test set. The training hyperparameters of the semantic segmentation models are the same as the semantic segmentation methods in the previous section, the training hyperparameters of the SAM based models are the same as the comparison of SAM with different prompt in the previous section, and the hyperparameters trained by YOLOX as the Prompt Generator in TongueSAM are the same as the object detection method in the previous section. The result is shown in Table 2.

The experiment indicate that, for conventional semantic segmentation methods, the performance on TongueSet1 surpasses that on TongueSet2 and TongueSet3. This can be attributed to the fact that TongueSet1 has the largest number of training samples and consistent capture environment. SAM based models outperform others on all three datasets. Even on challenging datasets such as TongueSet3, SAM based models achieves excellent segmentation results.

\subsection{Comparison with other tongue segmentation methods with zero-shot.}
The strength of TongueSAM lies in its zero-shot segmentation capability. After pretraining the models mentioned in the previous section on TongueSet1, this paper valids their performance on TongueSet2 and TongueSet3, respectively. The experimental results are presented in Table 3.

The experimental results indicate that TongueSAM exhibits a significant advantage outperforms other methods with zero-shot. Compared to SAM without prompts, TongueSAM closely approximates the SAM with ground truth prompts. This observation underscores the effectiveness of the Prompt Generator. To provide a visual comparison of the segmentation methods, this study randomly selected several tongue images from TongueSet2 and TongueSet3. The selected images are illustrated in Fig. 5.

According to the comparison of the different segmentation results, it can be found that the general model often produces wrong segmentation in the face of objects that do not appear in the training set. SAM, on the other hand, has been pretrained on a large number of nature images and will have a clear understanding of various types of objects, even if these objects do not appear in the tongue training set. At the same time, tongue detection also has good adaptability and robustness, which enables Prompt Generator to generate prompt similar to Ground Truth.

\section{Conclusion}

This paper presents TongueSAM, an universal tongue segmentation method based on the large-scale interactive segmentation model SAM. TongueSAM achieves superior segmentation results compared to other methods, particularly in zero-shot scenarios. TongueSAM demonstrates exceptional adaptability and robustness, enabling it to achieve optimal segmentation results even on challenging samples. In future work, we aim to enhance the efficiency of TongueSAM, enabling its flexible application in diverse scenarios.

\section{Acknowledge}
This work was supported by Industry-University-Research Cooperation Project of Fujian Science and Technology Planning (No:2022H6012),Natural Science Foundation of Fujian Province of China (No.2021J011169, No.2022J011224), Science and Technology Research Project of Jiangxi Provincial Department of Education (No.GJJ2206003). Special thanks to my friends Geng Dong, Zhizhong Zhang, and his wife Xue Wang for their encouragement.


\begin{thebibliography}{8}

\bibitem{1}Mahmoud Marhamati, Ali Asghar Latifi Zadeh, Masoud Mozhdehi Fard, Mohammad Arafat Hussain, Khalegh Jafarnezhad, Ahad Jafarnezhad, Mahdi Bakhtoor, Mohammad Momeny:LAIU-Net: A learning-to-augment incorporated robust U-Net for depressed humans' tongue segmentation. Displays 76: 102371 (2023)
\bibitem{2}Changen Zhou, Haoyi Fan, Zuoyong Li:Tonguenet: Accurate Localization and Segmentation for Tongue Images Using Deep Neural Networks. IEEE Access 7: 148779-148789 (2019)
\bibitem{3}Hui Tang, Bin Wang, Jun Zhou, Yongsheng Gao:DE-Net: Dilated Encoder Network for Automated Tongue Segmentation. ICPR 2020: 2575-2581
\bibitem{4}Bingqian Lin, Junwei Xle, Cuihua Li, Yanyun Qu:Deeptongue: Tongue Segmentation Via Resnet. ICASSP 2018: 1035-1039
\bibitem{5}Xinfeng Zhang, Haonan Bian, Yiheng Cai, Keye Zhang, Hui Li:An improved tongue image segmentation algorithm based on Deeplabv3+ framework. IET Image Process. 16(5): 1473-1485 (2022)
\bibitem{6}Wanqiang Cai, Bin Wang:DSE-Net: Deep Semantic Enhanced Network for Mobile Tongue Image Segmentation. ICONIP (7) 2022: 138-150
\bibitem{7}Lei Li, Zhiming Luo, Mengting Zhang, Yuanzheng Cai, Candong Li, Shaozi Li:An iterative transfer learning framework for cross-domain tongue segmentation. Concurr. Comput. Pract. Exp. 32(14) (2020)
\bibitem{8}Qingbin Zhuang, Senzhong Gan, Liangyu Zhang:Human-computer interaction based health diagnostics using ResNet34 for tongue image classification. Comput. Methods Programs Biomed. 226: 107096 (2022)
\bibitem{9}Ye Yuan, Wei Liao:Design and Implementation of the Traditional Chinese Medicine Constitution System Based on the Diagnosis of Tongue and Consultation. IEEE Access 9: 4266-4278 (2021)
\bibitem{10}Xiaohui Lin, Zhaochai Yu, Zuoyong Li, Weina Liu:Machine Learning Based Tongue Image Recognition for Diabetes Diagnosis. ML4CS (3) 2020: 474-484
\bibitem{11}Alexander Kirillov, Eric Mintun, Nikhila Ravi, Hanzi Mao, Chloé Rolland, Laura Gustafson, Tete Xiao, Spencer Whitehead, Alexander C. Berg, Wan-Yen Lo, Piotr Dollár, Ross B. Girshick:Segment Anything. CoRR abs/2304.02643 (2023)
\bibitem{12}Bo Pang, Kuanquan Wang, David Zhang, Fengmiao Zhang:On Automated Tongue Image Segmentation in Chinese Medicine. ICPR (1) 2002: 616-619
\bibitem{13}Wangmeng Zuo, Kuanquan Wang, David Zhang, Hongzhi Zhang:Combination of polar edge detection and active contour model for automated tongue segmentation. ICIG 2004: 270-273
\bibitem{14}Jianqiang Du, Yansheng Lu, Mingfeng Zhu, Kang Zhang, Chenghua Ding:A Novel Algorithm of Color Tongue Image Segmentation Based on HSI. BMEI (1) 2008: 733-737
\bibitem{15}Yichi Zhang, Rushi Jiao:How Segment Anything Model (SAM) Boost Medical Image Segmentation? CoRR abs/2305.03678 (2023)
\bibitem{16}Christian Mattjie de Oliveira, Luis Vinícius de Moura, Rafaela Cappelari Ravazio, Lucas Silveira Kupssinskü, Otávio Parraga, Marcelo Mussi Delucis, Rodrigo Coelho Barros:Zero-shot performance of the Segment Anything Model (SAM) in 2D medical imaging: A comprehensive evaluation and practical guidelines. CoRR abs/2305.00109 (2023)
\bibitem{17}Mingzhe Hu, Yuheng Li, Xiaofeng Yang:SkinSAM: Empowering Skin Cancer Segmentation with Segment Anything Model. CoRR abs/2304.13973 (2023)
\bibitem{18}Peilun Shi, Jianing Qiu, Sai Mu Dalike Abaxi, Hao Wei, Frank P.-W. Lo, Wu Yuan:Generalist Vision Foundation Models for Medical Imaging: A Case Study of Segment Anything Model on Zero-Shot Medical Segmentation. CoRR abs/2304.12637 (2023)
\bibitem{19}Maciej A. Mazurowski, Haoyu Dong, Hanxue Gu, Jichen Yang, Nicholas Konz, Yixin Zhang:
Segment Anything Model for Medical Image Analysis: an Experimental Study. CoRR abs/2304.10517 (2023)
\bibitem{20}Kaiming He, Xinlei Chen, Saining Xie, Yanghao Li, Piotr Dollar, and Ross Girshick. Masked autoencoders are scalable vision learners. CVPR, 2022.
\bibitem{21}Alec Radford, Jong Wook Kim, Chris Hallacy, Aditya Ramesh, Gabriel Goh, Sandhini Agarwal, Girish Sastry, Amanda Askell, Pamela Mishkin, Jack Clark, et al. Learning transferable visual models from natural language supervision. ICML, 2021.
\bibitem{22}Nicolas Carion, Francisco Massa, Gabriel Synnaeve, Nicolas Usunier, Alexander Kirillov, and Sergey Zagoruyko. End-to-end object detection with Transformers. ECCV, 2020.
\bibitem{23}Bowen Cheng, Alex Schwing, and Alexander Kirillov. Per- pixel classification is not all you need for semantic segmentation. NeurIPS, 2021.
\bibitem{24}https://github.com/BioHit/TongeImageDataset
\bibitem{25}https://aistudio.baidu.com/aistudio/datasetdetail/196398

\end{thebibliography}
\end{document}